\documentclass[letterpaper, 10 pt, conference]{ieeeconf}  % Comment this line out if you need a4paper

\IEEEoverridecommandlockouts                              % This command is only needed if 
\usepackage{cite}
\usepackage{ulem} 
\usepackage{balance}
\usepackage{graphicx} %% for loading jpg figures
\usepackage{stfloats}
\usepackage{hyperref}   % to set up hyperlinks
\usepackage{float}
\usepackage[ruled]{algorithm2e}%伪代码用的
\usepackage{caption}
\usepackage{multirow} % Required for multirows
\usepackage{booktabs}%绘制三线表格用的
\usepackage{indentfirst}%首行缩进的
\usepackage{subfigure}
\usepackage{color}  % 改字体颜色的
\usepackage{lipsum} % 输出随机文本
\usepackage{amsmath,esint} % 输出数学公式
\usepackage{cuted}
\usepackage{CJKutf8}
\usepackage{verbatim}
\usepackage{bm}
\usepackage{setspace}
\usepackage{array,caption}
\usepackage{fancyhdr}
\usepackage[labelfont=bf]{caption}
\usepackage{ulem}
\usepackage{color}
\usepackage{xcolor}
\usepackage{amsfonts}
\usepackage{tabularray}

\hypersetup{
	colorlinks=true,
	linkcolor=blue,
	citecolor=blue,
	urlcolor=blue,
}
\IEEEoverridecommandlockouts                              % This command is only needed if 
\title{\LARGE \bf
Map Imagination Like Blind Humans: Group Diffusion Model for Robotic Map Generation
}

\author{Qijin~Song, and Weibang~Bai$^{*}$
\thanks{
This work is supported by the Shanghai Pujiang Program (23PJ1408500), Shanghai Frontiers Science Center of Human-centered Artificial Intelligence (ShangHAI), MoE Key Laboratory of Intelligent Perception and Human-Machine Collaboration (KLIP-HuMaCo).
The experiments of this work were supported by the Core Facility Platform of Computer Science and Communication, SIST, ShanghaiTech University.
\textit{(Corresponding author: Weibang Bai)}
}% <-this % stops a space
\thanks{Qijin Song and Weibang Bai are with 
the ShanghaiTech Automation and Robotics (STAR) Center, School of Information Science and Technology, ShanghaiTech University, Shanghai, 201210, China. }
}

\begin{document}

\maketitle
\thispagestyle{empty}
\pagestyle{empty}
%%%%%%%%%%%%%%%%%%%%%%%%%%%%%%%%%%%%%%%%%%%%%%%%%%%%%%%%%%%%%%%%%%%%%%%%%%%%%%%%
\begin{abstract}
Can robots imagine or generate maps like humans do, especially when only limited information can be perceived like blind people? To address this challenging task, we propose a novel group diffusion model (GDM) based architecture for robots to generate point cloud maps with very limited input information.
Inspired from the blind humans' natural capability of imagining or generating mental maps, the proposed method can generate maps without visual perception data or depth data. With additional limited super-sparse spatial positioning data, like the extra contact-based positioning information the blind individuals can obtain, the map generation quality can be improved even more.
Experiments on public datasets are conducted, and the results indicate that our method can generate reasonable maps solely based on path data, and produce even more refined maps upon incorporating exiguous LiDAR data.
Compared to conventional mapping approaches, our novel method significantly mitigates sensor dependency, enabling the robots to imagine and generate elementary maps without heavy onboard sensory devices.
\end{abstract}

%%%%%%%%%%%%%%%%%%%%%%%%%%%%%%%%%%%%%%%%%%%%%%%%%%%%%%%%%%%%%%%%%%%%%%%%%%%%%%%%
\section{Introduction}
% \subsection{Pilosophy Concept}
Mapping is fundamental for navigation, planning, and efficient decision-making across all environments, as it helps to understand spatial relationships ~\cite{zhang2014loam,chang2007p}. 
Humans naturally tend to reconstruct the map of our surroundings to guide real-time movements, even with limited local observations.
For blind individuals, the amount of observed information is significantly reduced due to the lack of visual perception.
However, they can still form mental maps to guide movement and decision-making, relying heavily on virtual or mental odometry from memory, or imagination, as well as real or physical contact feedback with the environment through their body or walking sticks.

% \subsection{Technical Background}
In general, LiDAR scanning and visual sensing are widely used for robotic perception and mapping. Traditional approaches usually focus on feature matching to integrate multiple frames of data into a map~\cite{zhang2014loam, campos2021orb, qin2018vins}. 
% However, it is challenging to predict or generate a global map with limited information, such as relying solely on the path data and super-sparse LiDAR points~\cite{nunes2024scaling}. %goooooood----
However, global map prediction or generating is still challenging especially when only limited information can be obtained~\cite{nunes2024scaling}. It is, therefore, becoming increasingly important and popular in robotic mapping as well as in the rapidly growing autonomous driving industry. 

To generate maps, some approaches focus on using vehicle tracking data or pedestrian trajectories combined with neighborhood building footprints~\cite{davies2006scalable,yang2023mac}.
% \bluecolor{Some works focus on generating maps using vehicle tracking data~\cite{davies2006scalable} or pedestrian trajectories with neighborhood building footprints~\cite{yang2023mac}. 
By fusing historical information from multiple vehicles or pedestrians, a 2D route map can be generated, but producing a detailed 3D point cloud map remains challenging. 
Therefore, some approaches also try to create a 3D point cloud representation, either by synthesizing it from random noise~\cite{cheng2021learning} or by utilizing scanned LiDAR data points~\cite{lee2023diffusion, nunes2024scaling}. The process of generating such point clouds is not easy. 
Generative Adversarial Networks (GANs) and Variational Autoencoders (VAEs) are utilized to tackle this task~\cite{cheng2021learning, anvekar2022vg}, but these models exhibit limited capabilities in generating large-scale maps.

Recently, diffusion models, such as DDPM~\cite{ho2020denoising}, have been popularly used for image generation in the field of computer vision. Furthermore, DDIM~\cite{song2020denoising} proposed a new sampling method to speed up the generation process. LDM~\cite{rombach2022high} encodes images into a latent space and combines with multi-modal information to generate higher quality image data. 
In the meantime, diffusion models are also used to create LiDAR and large-scale scene point clouds. 
They are categorized into two main technical approaches:
One converts point clouds to range images, and use vision techniques to generate range image and then converte them back to point clouds, suitable for sensor-collected point clouds like LiDAR~\cite{ran2024towards, helgesen2024fast}. The other uses pointnet or 3D convolutions, suitable for scene-level point cloud generation~\cite{liu2024pyramiddiffusionfine3d}. These methods can efficiently generate 3D point cloud maps, but they need LiDAR scans or images as conditional input.

Inspired by the blind humans' ability to imagine and generate the global map from exiguous information, and amazed by the great potential of diffusion models' data generation ability, we are proposing a novel method enabling robots to predict large-scale maps based on very limited sensory singals such as only with the path data. We aim to generate a broader range of point cloud maps utilizing solely path data or with minimal extra point cloud information.
The main contributions can be summarized as follows:

\begin{itemize}
\item We propose a novel Group Diffusion Model (GDM) for large-scale map generation. GDM is a point-wise method, thus we firstly segment the large-scale point cloud into multiple groups and then apply the diffusion process and the denoising process separately to these group points.
\item We propose a two-step method for generating 3D point cloud maps. Stage 1 creates central points from path data and add noise to them. Stage 2 focuses on denoising the noisy map from stage 1. 
\item We tested our method on open datasets, demonstrating its ability to generate large-scale point cloud maps using only path data. The proposed method can be further enhanced with a small amount of additional super-sparse, or exiguous positioning data, such as just a few sampled LiDAR points.
\end{itemize}
\begin{figure*}[t]
    \centering
    \includegraphics[width=0.9\linewidth]{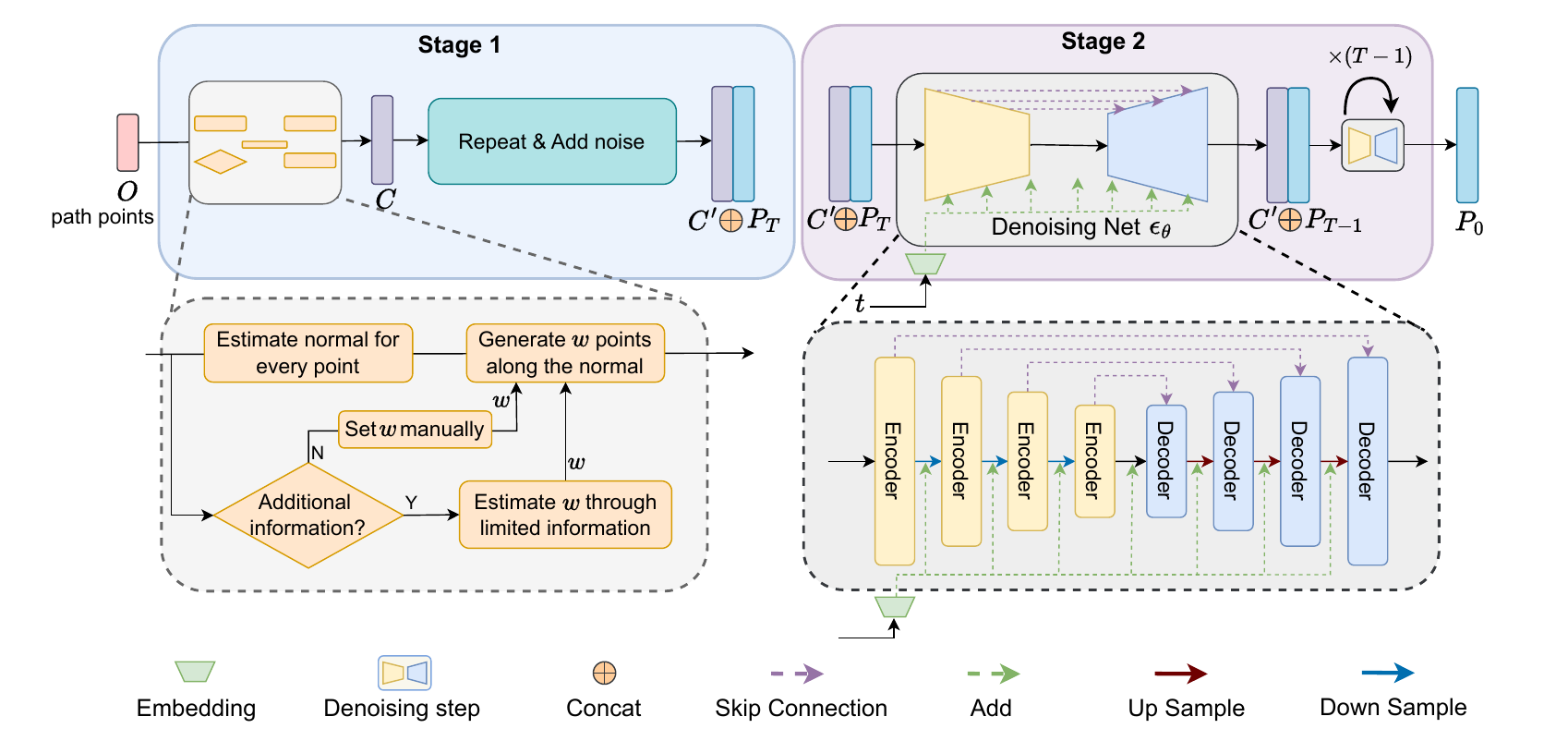}
     \caption{The architecture of our proposed two-stage map generation method. While stage 1 generate central points $C'$ and noisy map $P_T$, stage 2 employ the denoising process to generate large scale map $P_0$.}
     \label{fig:mapdiff}
     % \vspace{-0.25cm}
\end{figure*}

\section{Related Work}
Map generation aims to create a 3D point cloud map from random noise or scanned LiDAR data points~\cite{cheng2021learning,lee2023diffusion}. Previous work proposed many ways to generate unknown map, some of which leverage the geometric information of known maps to generate unknown maps~\cite{chang2007p, luperto2021completing}, while others employ neural network models, utilizing pre-training methods for generation~\cite{shrestha2019learned,katyal2019uncertainty,katyal2021high,sharma2024pre}. A similar way to generate unseen parts of a map is called scene completion, which tries to complete missing 3D details of a scene based on incomplete sensor data~\cite{nunes2024scaling}. 

Denoising diffusion probabilistic models have become popular because they generate high quality images and videos~\cite{ho2020denoising,nichol2021improved,khachatryan2023text2video,liang2024rich,wang2024instancediffusion}. One significant advantage of diffusion models is their ability to produce high-fidelity outputs. Compared to other generative models like GANs (Generative Adversarial Networks)~\cite{goodfellow2020generative}, diffusion models tend to generate samples with fewer artifacts and more intricate details, making them particularly suitable for tasks that require precise control over the generated content.Due to the time-consuming drawback of diffusion models, numerous studies have focused on improving their efficiency, reducing their computational requirements, and extending their capabilities to new domains. For example, some studies have proposed techniques to accelerate the sampling process, allowing diffusion models to generate images and videos faster~\cite{song2020denoising,ma2024deepcache}. Others have explored ways to condition the diffusion process on specific inputs, enabling the models to generate samples that meet specific criteria or follow certain styles~\cite{zhang2023adding,mo2024freecontrol}.

Diffusion model for LiDAR is difficult since its large scale. Some works transfer LiDAR scan to range image and employ image generation method to generate LiDAR scan~\cite{zyrianov2022learning,ran2024towards}. This method is not suitable for large scale map generation, since the map can not be transfered to range image. Some works propose point cloud diffusion to generate object point cloud~\cite{luo2021diffusion,vahdat2022lion}. For large scale map generation, the work~\cite{liu2023pyramid} employs scale-varied diffusion models to generate high-quality outdoor scenes. This work achieves the generation of large-scale maps by combining the generated small maps. 

Different from previous methods, our approach aims to generate large scale point cloud maps without using vision or LiDAR based perception, eliminating the necessity for onboard LiDAR or visual sensors.

\section{Methodology}

Firstly, inspired from the natural abilities of blind and deaf humans, robots should also be equiped with similar map generation intelligence utilizing limited sensory systems onboard. Without regard to the voice peception, the blind individuals are mainly relying on their path memories and some random contact feedback with the environment using hands and walking sticks.
In fact, the path memories can achieve mental odometry, and the interactive environmental contact provides distance or relative positioning information. 

Whereas, with necessary simple inputs, the basic odometry information of robots can be easily obtained through encoders, GPS, IMU, etc., the random but super-sparse positioning information can be acquired by cheap ultrasonic sensors. To test our methods on open datasets, we can gather the small amount of super-sparse positioning data by directly sampling the LiDAR scanning. 

In this regard, we propose a group diffusion model for generating large-scale point cloud maps. As illustrated in Fig.\ref{fig:mapdiff}, this proposed approach includes two main stages. Stage 1 aims to generate central points from the given path, and stage 2 employs method defined by Sec. \ref{sec:gdml} to generate large-scale map.

\begin{figure}[h]
\vspace{-2mm}
    \centering
    \includegraphics[width=1\linewidth]{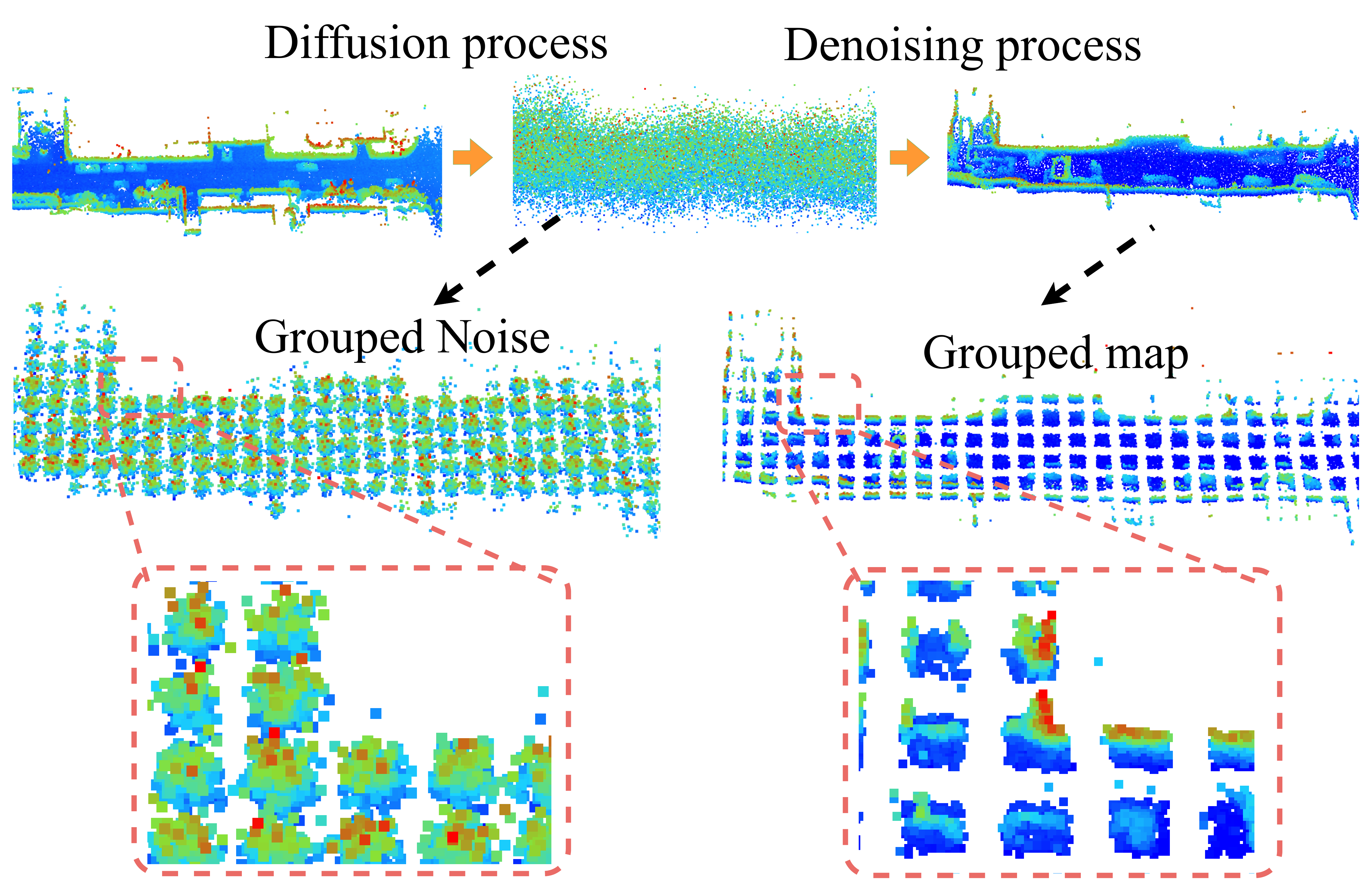}%groupintro3
     \caption{The group diffusion model works by dividing the original map into several groups. The diffusion process and the denoising process are separately applied to these group points. We add a certain amount of spacing between each group in this picture, while actual map is presented at the top of the picture without these spacing intervals.}
     \label{fig:groupintro}
     \vspace{-4mm}
\end{figure}

\subsection{Denoising Diffusion Probabilistic Models}
Denoising Diffusion Probabilistic Models (DDPM) ~\cite{ho2020denoising, song2020denoising} represent a generative modeling approach rooted in diffusion processes. These models generate data samples by progressively adding Gaussian noise to the data and learning to reverse this process. The methodology encompasses two primary phases: the diffusion process and the denoising process. During the diffusion process, noise is incrementally introduced to the data across $T$ steps, with the weighting of this noise typically determined by time-step-dependent beta coefficients that escalate with the progression of $t$. By the $T$-th step, the data is transformed into pure Gaussian noise. Conversely, the denoising process reverses this trajectory, gradually removing noise from the $T$-th step data to recover the original data. Typically, a neural network model is employed to predict the noise added at each step $t$, and by subtracting this noise, the data from the previous step is reconstructed. This iterative denoising process progressively approximates the original data.

The diffusion process constitutes a fixed Markov chain ~\cite{ho2020denoising}, systematically transforming the original data $x_0$ into noise-laden data $x_T$. Given the noise factors $\beta_t$, with $t=0,1,...,T$. Let $\alpha_t=1-\beta_t$,  each timestep $t$ of this transformation can be expressed as:
\begin{equation}
x_t = \sqrt{\bar{\alpha_t}}x_0
    + \sqrt{1-\bar{\alpha_t}}\epsilon
\end{equation}
with $\bar{\alpha_t}=\prod_{i=1}^{t}{\alpha_i}$, and $\epsilon$ is a Gaussian noise with mean 0 and the identity matrix $I$ as diagonal covariance.

The denoising process systematically eliminates noise at the $T$-th step, aiming to restore the original data. As per the definitions outlined in~\cite{ho2020denoising}, this denoising process can be formulated as:
\begin{equation}
x_{t-1}=\frac{1}{\sqrt{\alpha_t}}\left (x_t-\frac{\beta_t}{\sqrt{1-\bar{\alpha_t}}}\epsilon_{\theta}(x_t,t)\right )+\sqrt{\beta_t}z 
\end{equation}
where $z$ is the Gaussian noise, and $\epsilon_{\theta}(x_t,t)$ is the noise predicted from $x_t$ at timestep $t$. Finding a formula to predict noise is difficult, thus Ho \textit{et al.}~\cite{ho2020denoising}  define $\epsilon_{\theta}(x_t,t)$ as a neural network model.

The loss function is employed to optimize the neural network model $\epsilon_{\theta}(x_t,t)$. In one training step, given a random $t$ from $0$ to $T$ and origin data $x_0$, sample a Gaussian noise $\epsilon$ and calculate the t-th step noisy data $x_t$ using Eq.(1). Then, input $x_t$ and $t$ into the model to predict noise $\epsilon_{\theta}$. The loss function $L(\epsilon,t)$ can be formulated as:
\begin{equation}
L(\epsilon,t)=||\epsilon
    -\epsilon_{\theta}(x_t, t)||^2
\end{equation}

\subsection{Group Diffusion Model For LiDAR Map}
\label{sec:gdml}
% \begin{figure}[t]
%     \centering
%     \includegraphics[width=0.8\linewidth]{Figures/groupintro.pdf}
%      \caption{The group diffusion model.}
%      \label{fig:groupintro}
%      \vspace{-0.25cm}
% \end{figure}
The scale of LiDAR point cloud maps is often substantial, characterized by a vast range along the x-y axes and a comparatively narrow range along the z-axis, leading to a significant deviation in the distribution of point cloud data from the standard normal distribution. To address this issue, literature~\cite{nunes2024scaling} introduced a local diffusion model, which applies diffusion to individual points, proving effective within a single LiDAR frame but insufficient for larger-scale point cloud maps. Consequently, we propose a group diffusion model, where in the extensive point cloud is partitioned into numerous groups, and diffusion is individually applied to the point clouds within each group.

In Fig. \ref{fig:groupintro} the group noisy map does not represent the actual map we generate.To facilitate observation, we segregated each group by a certain distance. As observable, the group noisy map comprises multiple clusters of point clouds. By individually denoising each cluster, we obtain the group map, and integrating all these groups results in the generated map.

The diffusion process is different from DDPM ~\cite{ho2020denoising}. Given a ground truth of map $P$, and divide it into $m$ groups to get $p^i$ with $i=0,1,...,m$, that is, $P=\{p^0,p^1,...,p^m\}$. The distribution of a certain group is close to mean 0 and the identity matrix \textit{I} as diagonal covariance. Each group has a central point $C^i$:
\begin{equation}
C^i=\frac{1}{|p^i|}\sum_{x \in p^i}x
\end{equation}
where $|p^i|$ is the quantities of points in group $p^i$.
The central point $C^i$ is employed to get the $i-th$ normalized group $g^i=p^i-C^i$. Given a origin normalized group $g_0^i$, the diffusion process adds noise to $g_0^i$ over $T$ steps, resulting in $g_1^i,g_2^i,...,g_T^i$. From Eq. (1), the $t$-th step diffusion porocess can be rewritten as:
\begin{equation}
g_t^i= \sqrt{\bar{\alpha_t}}g^i_0
    + \sqrt{1-\bar{\alpha_t}}\epsilon, (i=0,1, \cdots, m; t \in [0, T])
\end{equation}
Transform $g_t^i$ into the coordinate system of $C^i$ to get the noisy group:
\begin{align}
p_t^i &= C^i + g_t^i \\
      &= C^i + \sqrt{\bar{\alpha_t}}g^i_0
       + \sqrt{1-\bar{\alpha_t}}\epsilon \\
      &= C^i + \sqrt{\bar{\alpha_t}}(p^i-C^i)
       + \sqrt{1-\bar{\alpha_t}}\epsilon
\end{align}
The complete noisy map can be represented as:
\begin{equation}
P_t=\{p^0_t, \ p^1_t, \cdots, \ p^m_t\}
\end{equation}

The denoising process involves the removal of noise from $p_t$ at the $t$-th step, resulting in data prior to the introduction of noise. Group diffusion necessitates denoising within individual groups. From Eq. (2), the deboising process can be rewritten as:
\begin{equation}
g_{t-1}^i=\frac{1}{\sqrt{\alpha_t}}\left (g_t^i-\frac{\beta_t}{\sqrt{1-\bar{\alpha_t}}}\epsilon_{\theta}(g_t^i,t)\right )+\sqrt{\beta_t}z
\end{equation}
Transform $g_{t-1}^i$ into the coordinate system of $c^i$ to get the group:
\begin{equation}
p_{t-1}^i=C^i + g_{t-1}^i
\end{equation}
Given a noisy map $P_t$, the denoising process is aim to remove noise at step $t$ and get the $P_{t-1}$:
\begin{equation}
P_{t-1}=\{p^0_{t-1},p^1_{t-1},...,p^m_{t-1}\}
\end{equation}

The loss function is employed to optimize the neural network model $\epsilon_{\theta}(g_t^i,t)$. In one training step, given a random $t$ from $0$ to $T$ and origin data $P_0$, sample a Gaussian noise $\epsilon$ and calculate the $t$-th step noisy map $P_t$ using Eq.(5-8). Then, input $P_t$ and $t$ into the model to predict noise $\epsilon_{\theta}$. The mean square error loss function $L_{mse}(P_t, t)$ can be formulated as:
\begin{equation}
L_{mse}(P_t, t)=\frac{1}{N}\sum_{i=0}^{m}||\epsilon-\epsilon_\theta(P_t,t)||^2
\end{equation}
The noise that we incorporate adheres to a standard normal distribution. Consequently, to accelerate training, we incorporate the regularization losses as mentioned in~\cite{nunes2024scaling}, resulting in the formulation of a comprehensive loss function. Given the mean $E(\epsilon_\theta)$ and the standard deviation $D(\epsilon_\theta)$. Computing $L_{mean}={\left(E(\epsilon_\theta)-0\right)}^2$ and $L_{std}={\left(D(\epsilon_\theta)-1\right)}^2$. The loss can be formulated as:
\begin{equation}
L=L_{mse}+r(L_{mean}+L_{std})
\end{equation}
where $r$ is a weighting factor, we set the default value of $r$ to 5.

Network Structure show in the right part of  Fig. \ref{fig:mapdiff}. We employ the sparse unet~\cite{pointcept2023} as our backbone. In our network, the input includes noisy map $P_t$, group center coordinates $C$ and timesteps $t$, and the output is predicted noise $\epsilon_\theta$.

\subsection{Two-Stage Map Prediction}
\label{sec:two_stage}
\begin{figure}[t]
    \centering
    \includegraphics[width=1\linewidth]{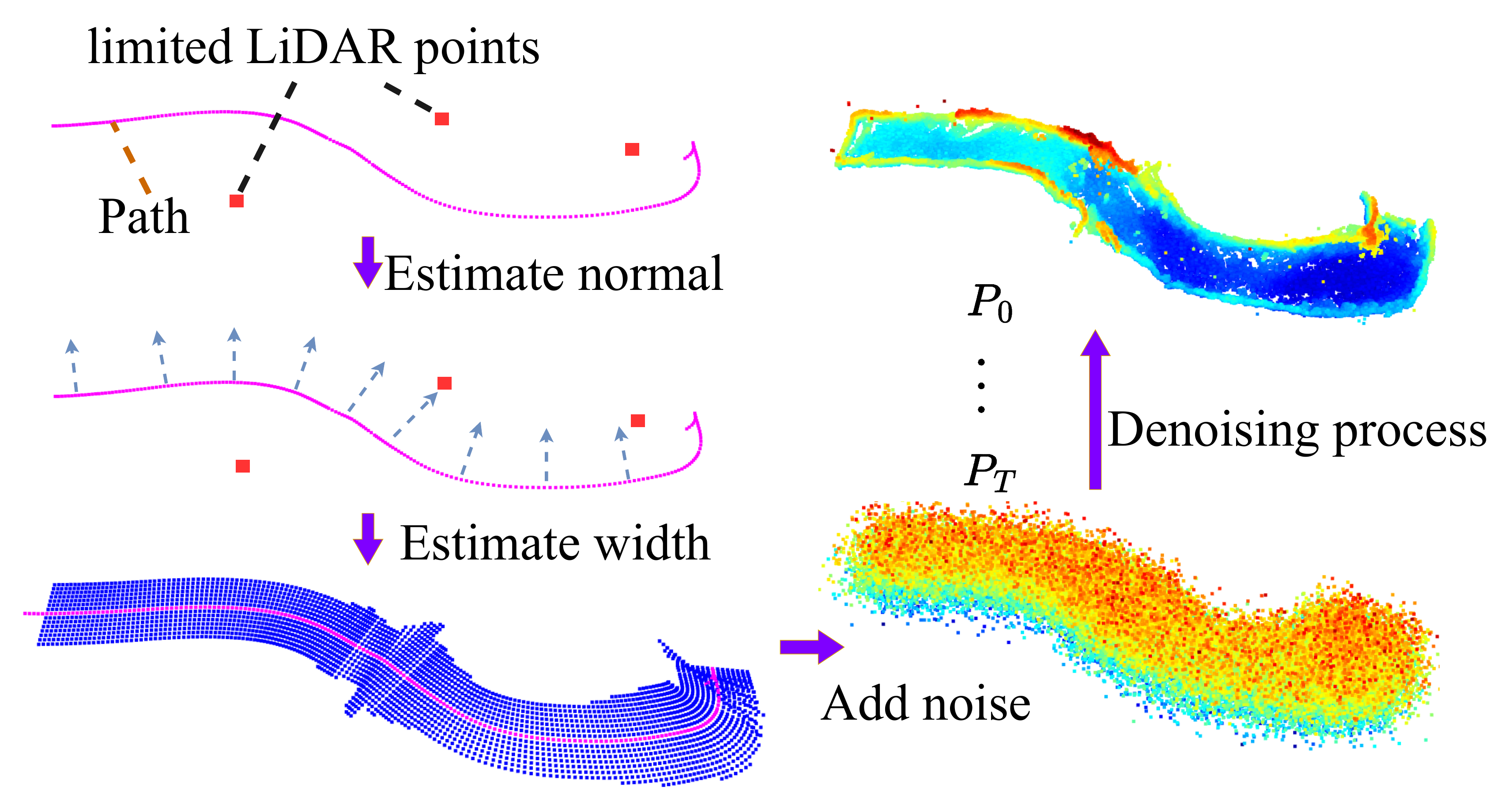}
     \caption{Generating map from a path and limited LiDAR points. Given both path data and limited LiDAR points, we first estimate their normals and width $w$, and then generate one point per meter along these normals, extending up to a distance of $w$ meters. Finally, we employ the diffusion process to add noise to the points in order to obtain $P_T$ and then we utilize the denoising process to obtain a detailed map $P_0$.}
     \label{fig:mapgen}
     \vspace{-5mm}
\end{figure}
Even with diffusion models, it remains challenging to generate large-scale map from path points, as we still require the $P_T$ and $C$ defined by Sec. \ref{sec:gdml}. The $P_T$  can be define as $P_T = C + \epsilon$ where $\epsilon$ is Guassian noise. To get $C$, we propose two-stage map prediction approach. Stage 1 aims to generate central points from path, and stage 2 employs method defined by Sec. \ref{sec:gdml} to predict large-scale map, as illustrated in Fig. \ref{fig:mapdiff}.

\textbf{Stage 1} aims to predict central points $C$. Our method is shown in the fist half of Fig. \ref{fig:mapdiff}. We create central points along the path, with a fixed width $w$. Given the path points $O=\{o_1,o_2,\cdots,o_i\}$ where $o_i \in \mathbb{R}^3$, we estimate normal vector $N=\{n_1,n_2,\cdots,n_{i}\}$ where $n_i \in \mathbb{R}^3$ for every point $o_i$, and generate one point per meter along the normal vector, up to a distance of $w$ meters. An alternative method is to generate central points with various widths using limited information, e.g., we can estimat the width through limited LiDAR data, as pictured in Fig. \ref{fig:mapgen}.
To estimate width $w$, we initially calculate the distance $d$ from the tangent at point on the path to the nearest neighbor LiDAR point, and then let $w = d$ as the estimated width.

\textbf{Stage 2} is the denoising process in Sec. \ref{sec:gdml}. Given a noisy map created by stage 1, the denoising process will eliminate noise at $T$ steps.

\section{Experiments and Discussion}
\textbf{Experimental datasets.} We train our model in KITTI-360~\cite{liao2022kitti} Datasets, which include sequences 00 to 10. Our method requires only a limited amount of data. Let the 00 sequence as traning data, and others as the test data. To process the data from KITTI-360~\cite{liao2022kitti}, we leverage the provided LiDAR and pose data to synthesize a block map every 150 meters. Subsequently, 50,000 points are sampled using FPS (Farthest Point Sampling)~\cite{qi2017pointnet++}, which serve as our ground truth for subsequent experiments. We processed the entire sequence 00, resulting in 604 block maps.

\textbf{Training.} We train our model on a NVIDIA RTX 3060 Ti GPU equipped with 16GB of memory, with a batch size of 1 and 200 epochs. The entire training duration was 24 hours, during which the GPU memory consumption peaked at 6GB. For inference, the GPU memory utilization approximated 3GB.

\textbf{Evaluation.} We utilize chamfer distance (CD) and intersection-over-union (IoU) as the evaluation metric to assess the similarity between the generated map and the ground truth map. Chamfer Distance (CD) is employed to evaluate the similarity in shape between our generated point cloud and the ground truth, while intersection-over-union (IoU) is utilized to assess the proportion of the ground truth occupied by our generated point cloud.

\begin{figure*}[t]
    \centering
    \includegraphics[width=0.9\linewidth]{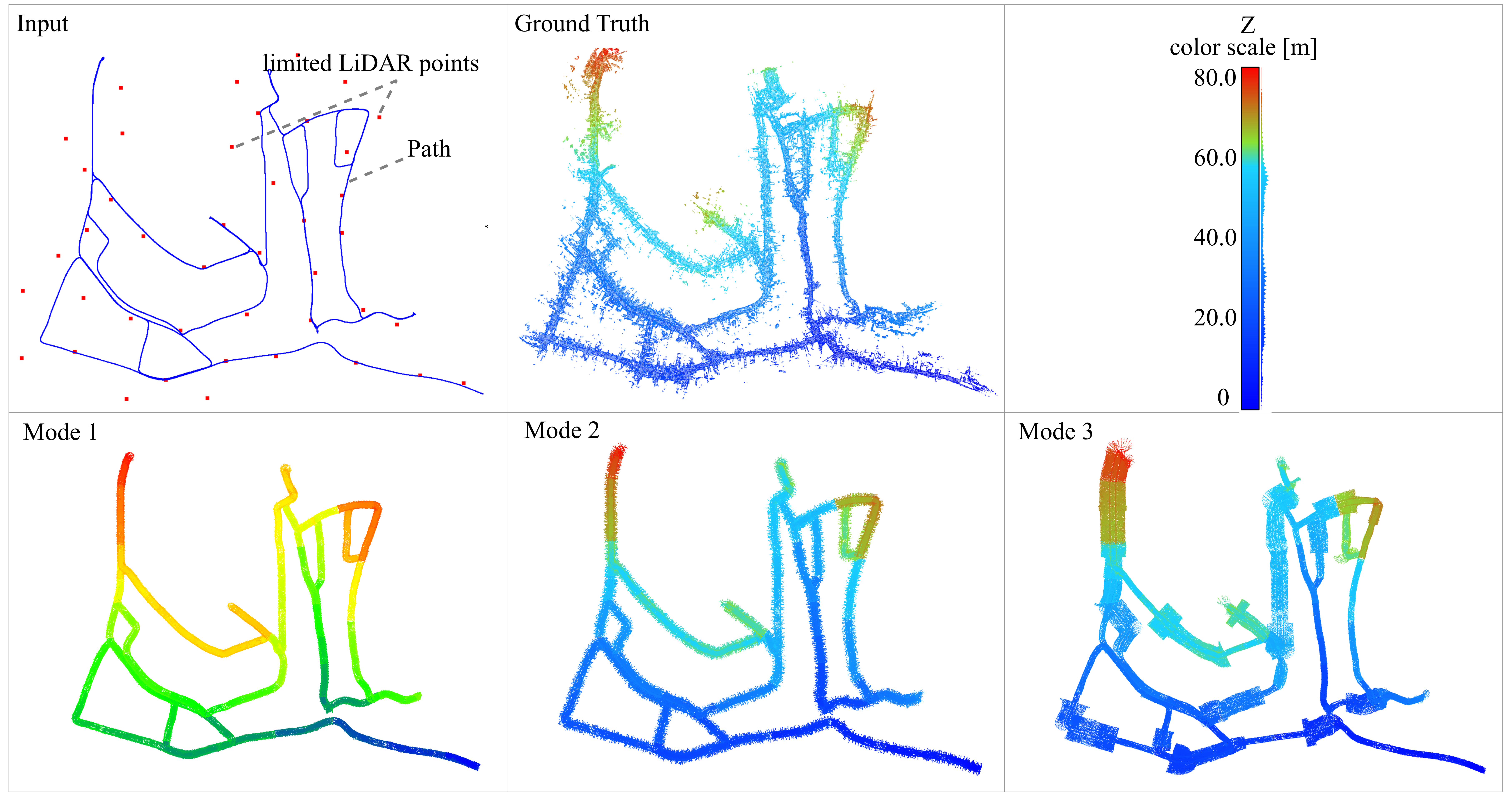}
    % \vspace{-1mm}
     \caption{Comparason of three map generation modes using our proposed two-stage map generation architecture. The length of this slected path is about 6.4km, and the height range is [0, 78m]. The key difference among the three modes lies in their input data: Mode 1 utilizes solely the path, Mode 2 incorporates both the path and a random width $w$, while Mode 3 employs the path alongside exiguous sampled LiDAR point clouds. }
     \label{fig:p2m_comp}
     \vspace{-4mm}
\end{figure*}

\begin{figure*}[b]
    \centering
    \vspace{-3mm}
    \includegraphics[width=0.95\linewidth]{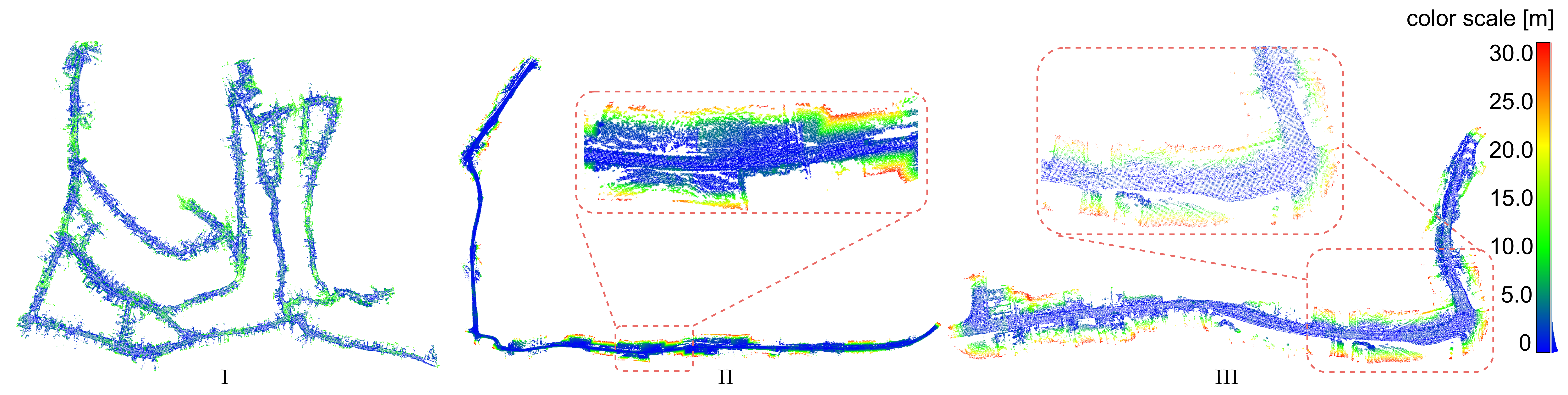}
    % \vspace{-1mm}
     \caption{The error distances color scale of the generated map compared to ground truth. We compare three types of map sequences (Seq) using the Mode 3. Seq I has the minimal error since it contains fewer outlier points.} % Seq II and Seq III exhibit higher errors due to having more outlier points compared to Seq I.}
     \label{fig:gt2m}
     % \vspace{-3mm}
\end{figure*}

\subsection{Map generation with path}
In this experiment, we create central points $C$ from the path data $O$. We process the poses file of KITTI-360~\cite{liao2022kitti} to get path $O=\{o_1,o_2,\cdots,o_i\}$ where $o_i \in \mathbb{R}^3$ is the point in path $O$. We employ the Stage 1 method proposed in Sec. \ref{sec:two_stage} to generate central point $C$, and set the width as a fixed value $w=20$. To analyze the generative effects as the width w varies, we devised Mode 2, which involves inputting path data and random width $w \in [15,35]$. In Fig. \ref{fig:p2m_comp}, we compare three modes in the dataset sequence I, whose path length is 6.4km and height range is [0, 78m]. The results indicate that, when utilizing path exclusively,  our method generates maps with a fixed width in Mode 1. In Mode 2, incorporating a random width results in the generation of increased details along the path.

\subsection{Map generation with path and limited point cloud}
In this experiment, we test the proposed Mode 3, which creates central points $C$ with the path data $O$ and limited point clouds sampled from the oringin map. To get the limited spatial positioning data like blind individuals, we here randomly sample only 50 points from the original map data, which has about $500,000$ points obtained from normal LiDAR scan. The method defined in Sec. \ref{sec:two_stage} is employed to estimate the width $w$ through the given limited LiDAR data.

Table~\ref{tab:eval} shows the results of three modes, where Mode 3 achieves the best performance in both metrics. Compared Mode 3 with Mode 1, we can conclude that, with the limited additional positioning information, the map generation metrics have improved by more than 20\%. The best performance of Mode 3 over the CD and IoU metric can be explained by the fact that Mode 3 make use of the limited LiDAR data, which enables it to produce a more detailed map compared
to other Mode. This is similar to a blind person acquiring additional information through touch or other means, thereby enabling them to imagine a more detailed map.
In Fig. \ref{fig:gt2m} we can compare the error between the modes. We can see that the Seq I has the minimal error since it contains fewer outlier points. Seq II and Seq III exhibit higher errors due to having more outlier points compared to Seq I.

\begin{table*}[t]
\centering
% \vspace{-3mm}
\caption{Statistical results of the map generation experiments. We testify our proposed method on 3 open datasets (I, II, and III) using three different map generating modes. The map generating Mode 1 uses only path data $O$ and fixed width $w=20 \text{m}$ as input. The map generating Mode 2 uses only path data and random width $w \in [15,35]$. The map generating Mode 3 uses both path data $O$ and exiguous sampled LiDAR data as input. Additionally, the specific width $w$ can be estimated through limited LiDAR data in Mode 3.} %Completion metric where the IoU is computed against the ground truth and generation grids with different resolutions.}
\label{table:g-Com}
% \vspace{-2mm}
% \resizebox{0.49\textwidth}{16mm}{
\begin{tabular}{cccccccc}
\hline\hline
\multirow{3}{1.1cm}{\begin{tabular}[c]{@{}c@{}}DataSet\\ Seq. \end{tabular}} & 
\multirow{3}{1.2cm}{\begin{tabular}[c]{@{}c@{}}Path\\ Distance\end{tabular}} &
\multirow{3}{1.1cm}{\begin{tabular}[c]{@{}c@{}}Map\\ Generating  \\ Mode \end{tabular}} &
\multirow{3}{1.5cm}{\begin{tabular}[c]{@{}c@{}}CD\\ Avg. [m]\end{tabular}} &
\multicolumn{3}{c}{\begin{tabular}[c]{@{}c@{}}IoU [\%] \\ Grid Resolution [m] \end{tabular}} & 
\multirow{3}{2cm}{\begin{tabular}[c]{@{}c@{}}Completion \\Time [s] \end{tabular}}\\
\cline{5-7} \\[-2mm]
  &  &  &  & \begin{tabular}[c]{@{}c@{}}  6m \end{tabular}  & \begin{tabular}[c]{@{}c@{}}  4m \end{tabular}  & \begin{tabular}[c]{@{}c@{}} 2m \end{tabular} &   \\[2mm]
 \hline \\ [-2mm]
% \midrule
I & 6.4 km & \begin{tabular}[c]{@{}c@{}} 1 \\ 2 \\ 3 \end{tabular}  
& \begin{tabular}[c]{@{}c@{}} 3.5  \\ 3.3  \\ \textbf{2.4}  \end{tabular} 
& \begin{tabular}[c]{@{}c@{}} 61.3 \\ 64.5 \\ \textbf{66.3} \end{tabular} 
& \begin{tabular}[c]{@{}c@{}} 50.5 \\ 51.9 \\ \textbf{53.8} \end{tabular} 
& \begin{tabular}[c]{@{}c@{}} 36.6 \\ 37.4 \\ \textbf{38.9} \end{tabular} 
& \begin{tabular}[c]{@{}c@{}} 886  \\ 884  \\           878 \end{tabular} \\
\midrule
II & 4.7 km & \begin{tabular}[c]{@{}c@{}} 1 \\ 2 \\ 3 \end{tabular} 
& \begin{tabular}[c]{@{}c@{}} 6.8  \\ 5.9  \\ \textbf{3.5}  \end{tabular} 
& \begin{tabular}[c]{@{}c@{}} 47.5 \\ 49.9 \\ \textbf{63.0} \end{tabular} 
& \begin{tabular}[c]{@{}c@{}} 38.2 \\ 40.8 \\ \textbf{52.9} \end{tabular} 
& \begin{tabular}[c]{@{}c@{}} 24.8 \\ 23.7 \\ \textbf{30.9} \end{tabular} 
& \begin{tabular}[c]{@{}c@{}} 565  \\ 570  \\           581 \end{tabular} \\
\midrule
III & 1.2 km & \begin{tabular}[c]{@{}c@{}} 1 \\ 2 \\ 3 \end{tabular} 
& \begin{tabular}[c]{@{}c@{}} 5.2  \\ 4.7  \\ \textbf{3.6}  \end{tabular} 
& \begin{tabular}[c]{@{}c@{}} 42.4 \\ 45.3 \\ \textbf{51.2} \end{tabular} 
& \begin{tabular}[c]{@{}c@{}} 34.0 \\ 39.2 \\ \textbf{48.5} \end{tabular} 
& \begin{tabular}[c]{@{}c@{}} 17.0 \\ 23.5 \\ \textbf{28.9} \end{tabular} 
& \begin{tabular}[c]{@{}c@{}} 176  \\ 159  \\          179  \end{tabular} \\
\hline\hline
\end{tabular}%
% }
\vspace{-1mm}
\label{tab:eval}
\end{table*}

\subsection{Map generation through any shape}
\begin{figure}[h]
    \centering
    \includegraphics[width=0.9\linewidth]{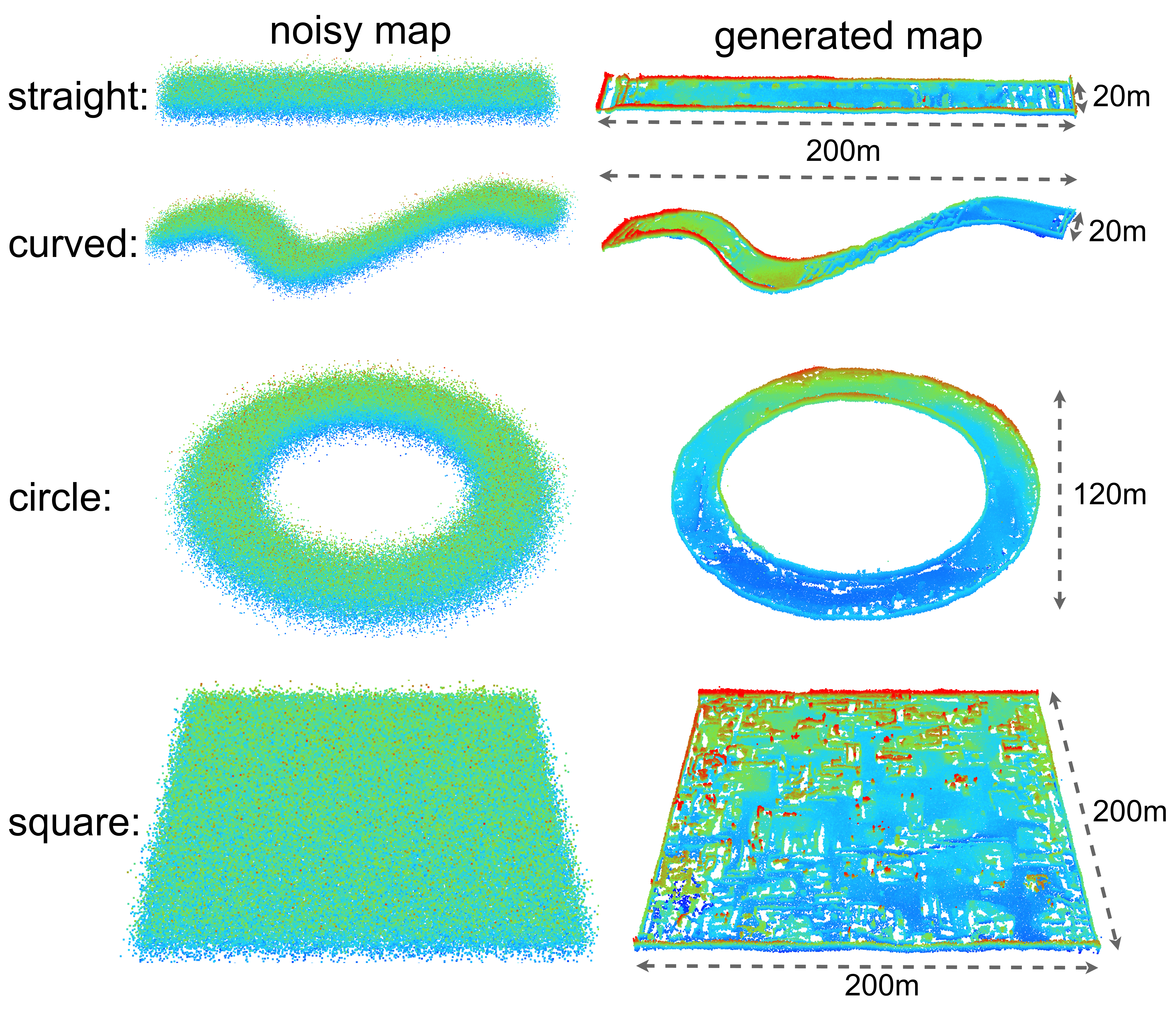}
     \caption{Four selected noisy map types are generated: a straight line with 200m in length and 20m in width, a curved version of the first one, a circular ring with a 120m diameter, and a 200m$\times$200m square-shaped map.}
     \label{fig:anymap}
     \vspace{-4mm}
\end{figure}

In this experiment, We manually constructed noisy maps of various shapes and generated the maps through the denoising process. The result is shown in Fig. \ref{fig:anymap}. Given a noisy map with any shapes, we employ the denoising process to get a map without noise. To create a noisy map, we initially establish its approximate outline, subsequently fill the map with a point cloud at 1-meter intervals, and add Gaussian noise to the point cloud. We manually created four noisy map types: a straight line 200m long and 20m wide, a curved version of this with bends, a circular ring 120m in diameter, and a square map 200$\times$200m. This experiment shows our method's ability to generate large maps of different shapes.

\section{Conclusion}
In this paper, we propose a novel method to generate large-scale point cloud map from only path data. To address the challenge of generating task, we propose the Group Diffusion method. The proposed map generation method includes two stages: In stage 1, we create central points from the path and add noise to them to get a noisy map. In stage 2, we employ a denoising process to generate a refined map. 
Experiments on public datasets showed that our method can generate reasonable maps using only path and refined maps with exiguous sampled LiDAR points. Three map generating modes are desgined and tested. When comparing Mode 3 with Mode 1, we can conclude that, with the limited additional positioning information, the map generation metrics have improved by more than 20\%. Compared to traditional approaches, our novel method reduces sensor dependency, enabling robots to create basic maps with minimal infomation, akin to blind humans relying on path memeory based mental odometry. Thus, robots acquire basic mapping abilities solely with odometry, reducing the need for LiDAR or vision sensors.

%\clearpage
% \bibliographystyle{unsrt}
% \bibliography{References} 

\bibliographystyle{IEEEtran}
\normalem
\balance
\bibliography{IEEEabrv, References}

\end{document}